\documentclass[10pt,twocolumn,letterpaper]{article}

\usepackage{cvpr}
\usepackage{times}
\usepackage{epsfig}
\usepackage{graphicx}
\usepackage{amsmath}
\usepackage{amssymb}
\usepackage{color}
\usepackage[labelsep=period]{caption}
\usepackage{subfigure}
\usepackage{caption}
\usepackage{cite}
\usepackage[ruled,vlined]{algorithm2e}
\usepackage{multirow}

\usepackage{url}
\usepackage[toc,page]{appendix}
\usepackage{pifont}
\usepackage[pagebackref=true,breaklinks=true,letterpaper=true,colorlinks,bookmarks=false]{hyperref}
\usepackage{array}
\usepackage{enumitem}

\newcommand{\tb}[1]{\textcolor{blue}{#1}}

\newcommand{\tr}[1]{\textcolor{red}{#1}}

\newcommand{\Cgray}[1]{\cellcolor[HTML]{D8D6D6}} %

\def\cvprPaperID{2743} 

\cvprfinalcopy

\begin{document}
	
	\title{Weighted Intersection over Union (wIoU) for Evaluating Image Segmentation}
	
	\author{\qquad Yeong-Jun Cho\\\qquad Department of Artificial Intelligence Convergence, Chonnam National Universiy\\{\qquad\tt\small yj.cho@chonnam.ac.kr}
		\and
	}
	
	\maketitle
	
	\begin{abstract}
		
		In recent years, many semantic segmentation methods have been proposed to predict label of pixels in the scene.
		In general, we measure area prediction errors or boundary prediction errors for comparing methods. 
		However, there is no intuitive evaluation metric that evaluates both aspects.
		In this work, we propose a new evaluation measure called weighted Intersection over Union (wIoU) for semantic segmentation. First, it builds a weight map generated from a boundary distance map, allowing weighted evaluation for each pixel based on a boundary importance factor. 
		The proposed wIoU can evaluate both contour and region by setting a boundary importance factor. 
		We validated the effectiveness of wIoU on a dataset of 33 scenes and demonstrated its flexibility.
		Using the proposed metric, we expect more flexible and intuitive evaluation in semantic segmentation field are possible.

	\end{abstract}
	

		\section{Introduction}
	
\begin{figure}[t]
		\centering
		\subfigure[Segmentation result~\cite{zhu2019improving} with an original image]{\includegraphics[width=0.95\columnwidth]{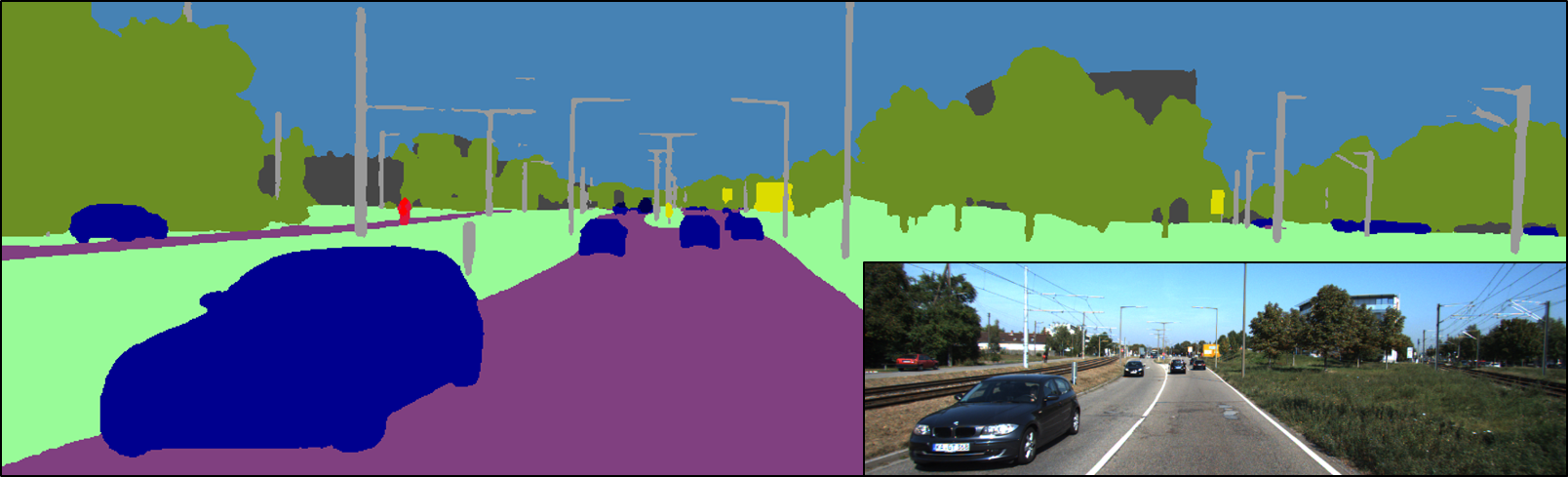}}
		\subfigure[Error map. Color of each pixel denotes: (green) True positive, (red) False negative, (yellow) False positive, (black) No label. ]{\includegraphics[width=0.95\columnwidth]{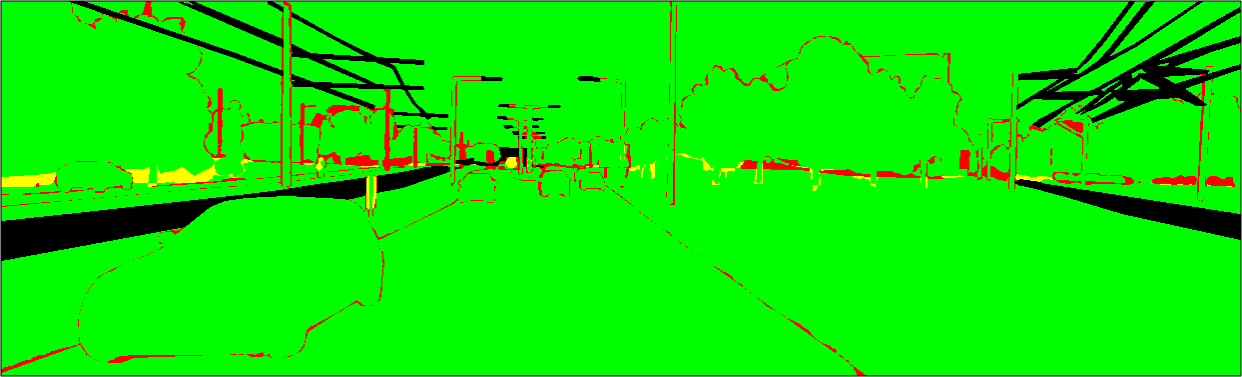}}
		\subfigure[Proposed weight map for the segmenation evaluation]{\includegraphics[width=0.95\columnwidth]{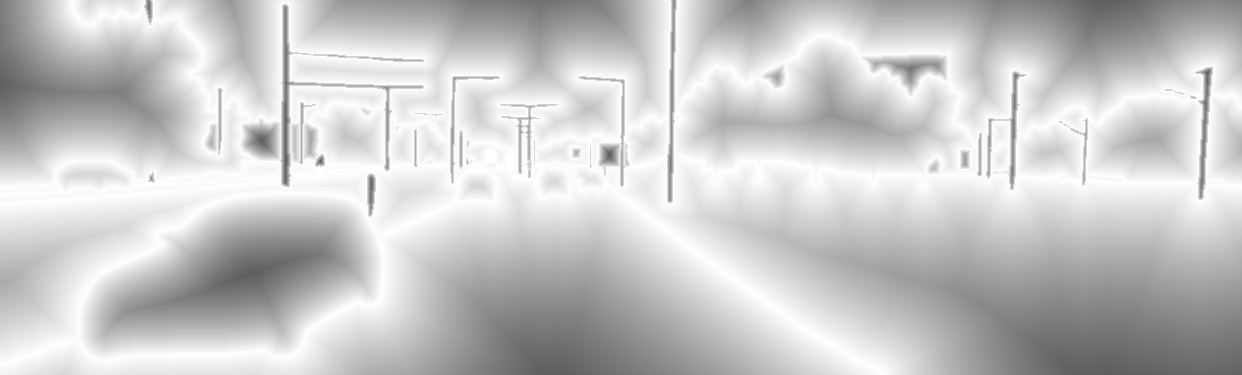}}
		\caption{Challenges in semantic segmentation and the proposed weight map. (a--b) It is difficult to predict pixel labels around object boundaries. (c) The proposed weight map emphasizes the importance of boundaries.}
		\label{FIG:challenges}
	\end{figure}

	In recent years, many studies have tried to train pixel-level classifiers on large-scale image datasets to perform accurate semantic image segmentation~\cite{zhu2019improving,takikawa2019gated}.  
	The goal of the image segmentation is to assign a class label of each pixel in the scene.
	It is one of the most basic problems in computer vision fields and has various potential applications e.g., autonomous driving systems~\cite{deng2019restricted}, scene understanding~\cite{malinowski2014multi} and medical image diagnostics~\cite{kayalibay2017cnn, hesamian2019deep}.
	Recently, significant advances in deep learning techniques are driving high-performance of semantic segmentation in various benchmark datasets~\cite{ everingham2010pascal,cordts2016cityscapes,Geiger2012CVPR}.
	Figure~\ref{FIG:challenges} shows a result of the semantic segmentation~\cite{zhu2019improving} and its error map in the \texttt{KITTI} dataset~\cite{Geiger2012CVPR}.

	Although the method performs the best compared to other state-of-the-art methods, it still experiences classification errors. Most of the errors~(i.e., false negatives and positives) occur around object boundaries due to high classification ambiguities around the boundaries as shown in Fig.~\ref{FIG:challenges}~(b).
	That is because at least two classes are adjacent around the objects' boundaries making class predictions more difficult and ambiguous.
	Even though the proportion of boundaries in the scene is very small, it has a significant impact on qualitative results. 
	In contrast, the internal areas of objects have relatively low classification ambiguities and occupy most of the scene area.
	
	In recent years, researchers have developed various methods to leverage boundary cues, such as adding additional network branches to baseline networks and optimizing boundary loss terms~\cite{takikawa2019gated, Bertasius_2016_CVPR, marmanis2018classification} due to the importance of boundary information.
	These methods aim to refine the precision of boundary detection, which in turn improves overall classification performance.
	To evaluate semantic segmentation results, most methods~\cite{zhu2019improving, takikawa2019gated} rank their methods by measuring intersection-over-union~(\textit{IoU}) score described in Section.~\ref{section:metrics}.
	
	{Unfortunately, \textit{IoU} evaluates all pixels equally, so the performance scores of many newly proposed methods improve very subtly according to the evaluation criterion.}
	To directly evaluate the boundary estimation results, several evaluation metrics such as BF score~\cite{csurka2013good} and Boundary Jaccard~(BJ)~\cite{fernandez2018new} were proposed; 
	however, they do not evaluate internal areas of the objects.
	In addition, those methods are highly dependent on the setting of distance tolerance parameter $\theta$.
	As we mentioned above, both evaluations of internal and boundaries of objects are important, without relying on certain tuning parameters.
	
	In this work, we propose a novel evaluation measure for semantic segmentation called weighted Intersection over Union~(\textit{wIoU}).
	It generates a weight map based on the boundary distance map, as shown in Figure.~\ref{FIG:challenges} (c). The weight map is adjustable  based on a boundary importance factor $\alpha$, as shown in Figure.~\ref{FIG:maps}. Using the weight map enables weighted evaluation for each pixel. 
	Setting the value of $\alpha$ is intuitive. Furthermore, it effectively evaluates both boundary and internal regions simultaneously.
	{
		Our work makes significant contributions to the field of image segmentation by introducing the following advancements:
		\begin{itemize}
			\item We introduce a novel segmentation evaluation metric called Weighted Intersection over Union (\textit{wIoU}), which provides a more comprehensive evaluation by integrating both the contour and region of the segments.
			\item Our proposed metric utilizes simple evaluation criteria, enabling it to assess segmentation performance effectively and intuitively.
			\item Unlike other metrics, \textit{wIoU} relies on minimal hyperparameters, with only one intuitive parameter: the boundary importance factor.
			\item We have validated the effectiveness of our proposed metric across a diverse set of test images, demonstrating its robustness and reliability in real-world scenarios.
		\end{itemize}
	}
	
	For comparing evaluation metrics, we construct a simple semantic segmentation dataset comprising 33 scenes.
	We have validated that the proposed \textit{wIoU} can evaluate both contours and regions according to the boundary importance factor $\alpha$.
	When the $\alpha$ value is small, it functions similarly to region-based evaluation metrics, particularly \textit{IoU}; otherwise it functions similar to boundary-based evaluation metrics.
	We anticipate that the proposed \textit{wIoU} will be used not only for evaluating metric but also as a novel region-contour loss for training accurate semantic segmentation networks.

	\begin{figure}[t]
		\centering
		\includegraphics[width=1\columnwidth]{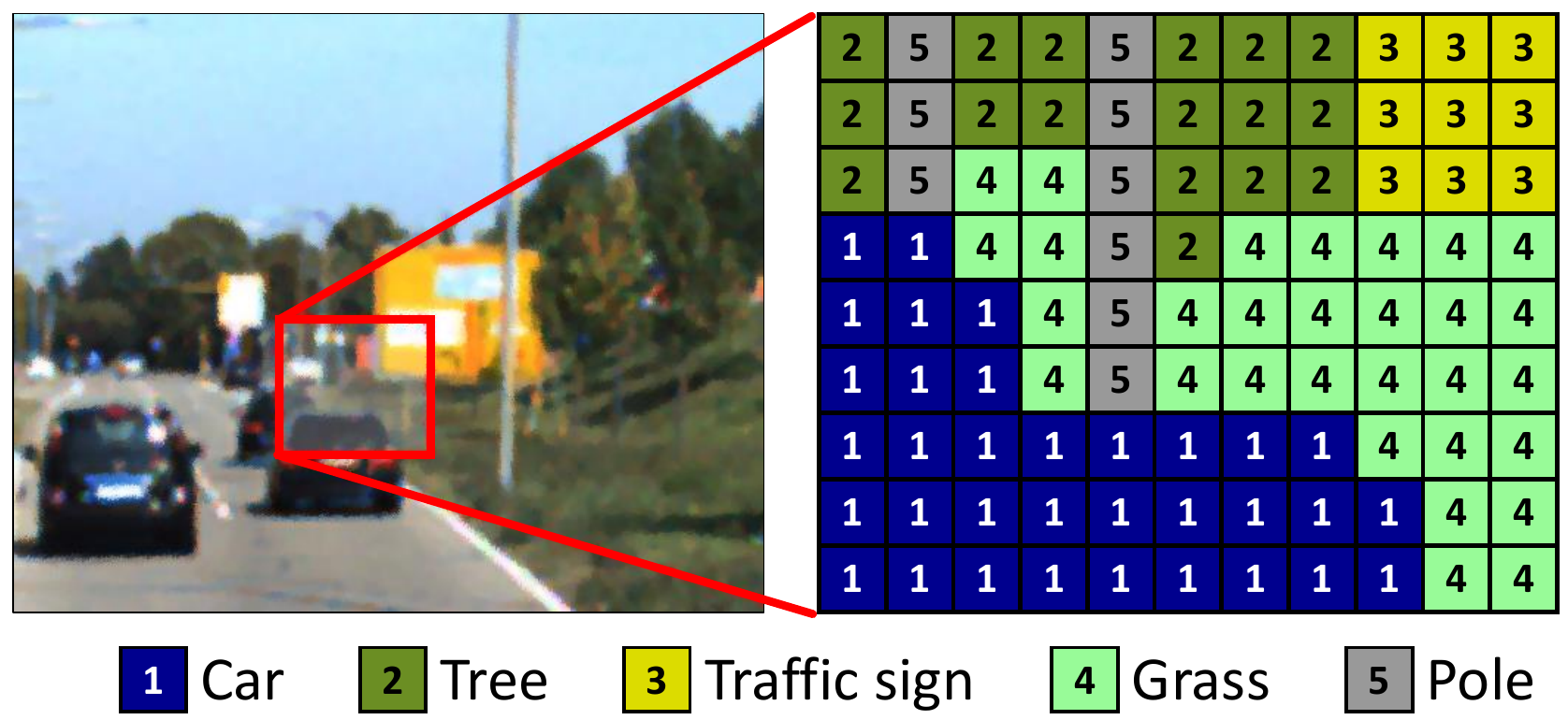}
		\caption{Example of an image and its predicted classes}
		\label{FIG:problem_def}
	\end{figure}

	\section{Problem Definition}
	\label{prob_def}
	Image segmentation~(also called semantic image segmentation) is a problem to find several sub-regions or partitions within the given image in the fields of computer vision and digital image processing. The sub-regions consist of sets of pixels belonging to objects.
	The goal of image segmentation is to predict the class labels~$\mathbf{C}$ of pixels in the given image~$\mathbf{I}$.
	An image can be represented as a set of pixels as follows,
	\begin{equation}
		\mathbf{I} = \left\{ { { i }_{ p } }| p = (p_x,p_y), 1{ \le  }p_x{ \le  }\mathrm{W},   1{ \le  }p_y{ \le  }\mathrm{H} \right\},
	\end{equation}
	where $i_p$ denotes the value of pixel $p$. $\mathrm{W}$ and $\mathrm{H}$ represent the width and height of the image, respectively.
	In general, an image consists of 3-channel (e.g., R,G,B), but for convenience, we do not consider the channel index.
	
	Using the given image pixels~${ i }_{ p }$, the probability distribution of class for each pixel is predicted as $p\left( { c }|{ { i }_{ p } } \right)$.
	To predict the probability distribution, we can employ various semantic segmentation approaches~\cite{zhu2019improving, takikawa2019gated}.
	Next, we select the class label for each pixel based on the distributions as follows,
	\begin{equation}
		{c}^{*}_{p} =  \underset { c\in \mathcal{C} }{ \text{argmax} }  p\left( { c }|{ { i }_{ p } } \right),
	\end{equation}
	where $\mathcal{C}$ is a set of class (e.g., car, tree, road and Etc).
	Finally, we build a predicted class map of the given image as
	\begin{equation}
		\mathbf{C} = \left\{ { {c}^{*}_{ p } }| p = (p_x,p_y), 1{ \le  }p_x{ \le  }\mathrm{W},   1{ \le  }p_y{ \le  }\mathrm{H} \right\}.
	\end{equation}
	Then, $\mathbf{C}\left( p \right)$ denotes the class label $c$ at the point $p=\left(p_x,p_y \right)$.
	Figure~\ref{FIG:problem_def} shows an example of a semantic segmentation result with predicted classes. As a result of semantic segmentation, each pixel corresponds to a distinct class label.

	\section{Related Works}

	\subsection{Semantic Image Segmentation Methods}
	Semantic image segmentation is one of the most important problems in the field of computer vision research.
	With the recent development of deep learning, many attempts to apply deep learning to semantic segmentation applications have been presented. 
	R.Girshick \textit{et al.}~\cite{girshick2014rich} proposed region proposal convolutional neural networks called R-CNN for object detection and semantic image segmentation. 
	In addition, J. Long \textit{et al.}~\cite{long2015fully} also proposed a semantic segmentation method via fully convolutional networks~(FCN) which is an end-to-end framework performing pixelwise prediction.
	Thanks to the large model capacity of deep neural networks, they significantly improved the prediction performance compared to many popular traditional methods~\cite{felzenszwalb2004efficient,roerdink2000watershed,carreira2010constrained, shotton2006textonboost,ladicky2009associative} reviewed in~\cite{thoma2016survey}.

	Recently, many methods~\cite{marmanis2018classification,yuan2020segfix,cheng2017fusionnet,liu2018ern, takikawa2019gated, Bertasius_2016_CVPR} have focused on the importance of boundary information in the scene to achieve accurate semantic segmentation.
	They added additional network branches and optimized the boundary loss to capture the boundary information.
	Then, the estimated boundary cues in the scene are effectively used for semantic segmentation.
	Many authors have argued that the boundary cue alleviates the problem of boundary blur and leads to edge-preserving results.
	Various benchmark datasets such as PASCAL VOC~\cite{everingham2010pascal}, Cityscapes~\cite{cordts2016cityscapes}, ADE20K~\cite{zhou2017scene}, ISPRS2D~\cite{ISPRS2D2018Semantic} and KITTI~\cite{Alhaija2018IJCV, Geiger2012CVPR} are commonly used to train and evaluate the methods.

	\subsection{Evaluation Metrics}
	\label{section:metrics}
	
	\begin{table}[t]
		\centering
		\begin{tabular}{cc|c|c|}
			\cline{3-4}
			&          & \multicolumn{2}{c|}{Predicted class} \\ \cline{3-4} 
			&          & Positive          & Negative         \\ \hline
			\multicolumn{1}{|c|}{\multirow{2}{*}{\begin{tabular}[c]{@{}c@{}}Actual\\ class\end{tabular}}} & Positive & \tb{TP}                & \tr{FN}               \\ \cline{2-4} 
			\multicolumn{1}{|c|}{}                                                                        & Negative & \tr{FP}                & \tb{TN}               \\ \hline
		\end{tabular}
		\caption{Confusion matrix for classification. Blue color indicates correct estimation and red color indicates failure estimation.}
		\label{tab01}
	\end{table}
	
	\noindent $\bullet$ \textbf{Basic notations for classification results.} \quad
	To evaluate the results of classification, there are four basic notations according to the actual class and predicted class:
	(1) True Positive~(TP); (2) True Negative~(TN); (3) False Positive~(FP); (4) False Negative~(FN). Each notation is summarized in Table.~\ref{tab01}. TP and TN are correct estimations. FP and FN are false alarms, which are called Type I error and Type II error, respectively.

	\noindent $\bullet$ \textbf{Region-based evaluation.} \quad	Based on the types of classification results, many studies in fields such as machine learning and pattern recognition have evaluated their methods.
	For example, the evaluation metrics such as Precision and Recall are very widely used:
	$Precision =  \frac{TP}{TP+FP},$ $Recall =  \frac{TP}{TP+FN},$
	Furthermore, $F_1$ score, which is the weighted average of \textit{Precision} and \textit{Recall}, has been used as an evaluation metric:
	\begin{equation}
		F_1 =  \frac{2TP}{2TP+FN+FP},  
		\label{eq:04}
	\end{equation}
	It reflects two evaluation results~(\textit{Precision} and \textit{Recall}) evenly.
	Another common metric for evaluating classification is a Jaccard index~(\textit{JI}):
	\begin{equation}
		JI =  \frac{TP}{TP+FN+FP},
	\end{equation}
	For semantic image segmentation problem defined in Section.~\ref{prob_def}, Intersection over Union~(\textit{IoU}) is equivalent to \textit{JI}:
	\begin{equation}
		IoU = JI = \frac { \left| \mathbf{C}\cap \mathbf{ C }_{ gt } \right|  }{ \left| \mathbf{C}\cup \mathbf{ C }_{ gt } \right|  }, 
		\label{eq:iou}
	\end{equation}
	where $\mathbf{C}$ denotes a predicted class map and $\mathbf{C}_{gt}$ denotes a ground-truth class label map.
	Most of the previous studies related to semantic segmentation~\cite{zhu2019improving, takikawa2019gated} have adopted \textit{IoU} measure.
	It compares a similarity between two regions~($\mathbf{C}$ and $\mathbf{C}_{gt}$) to evaluate all predicted pixels of the image.

	\begin{figure}[t]
		\centering
		\subfigure[]{\includegraphics[width=0.31\columnwidth]{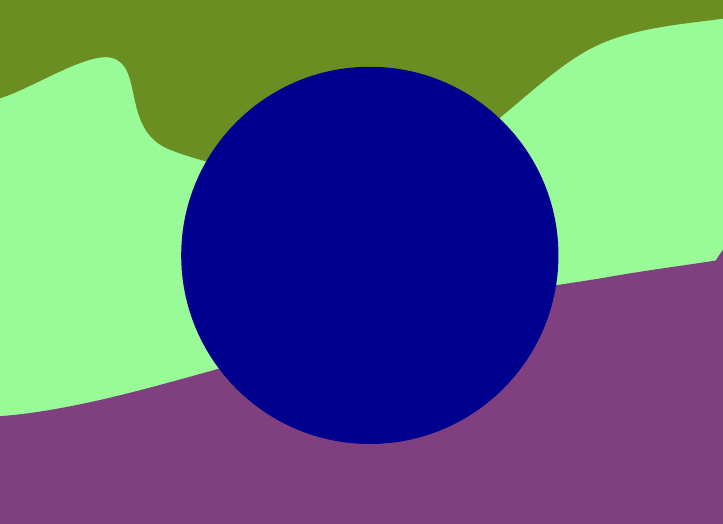}}\hspace{3pt}
		\subfigure[]{\includegraphics[width=0.31\columnwidth]{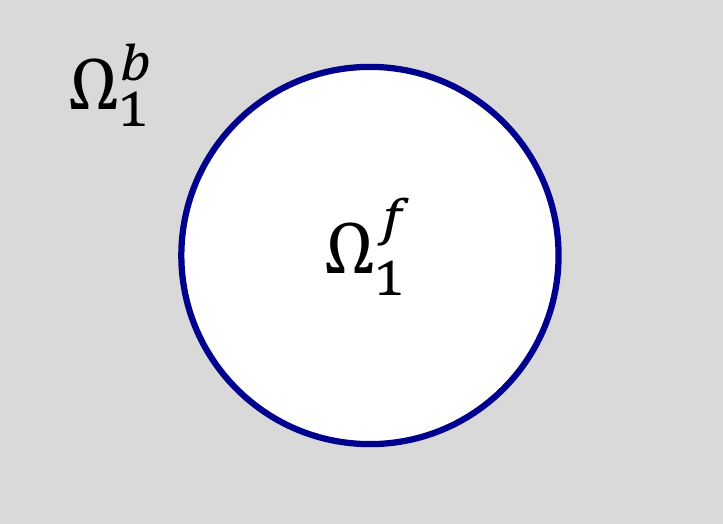}}\hspace{3pt}
		\subfigure[]{\includegraphics[width=0.31\columnwidth]{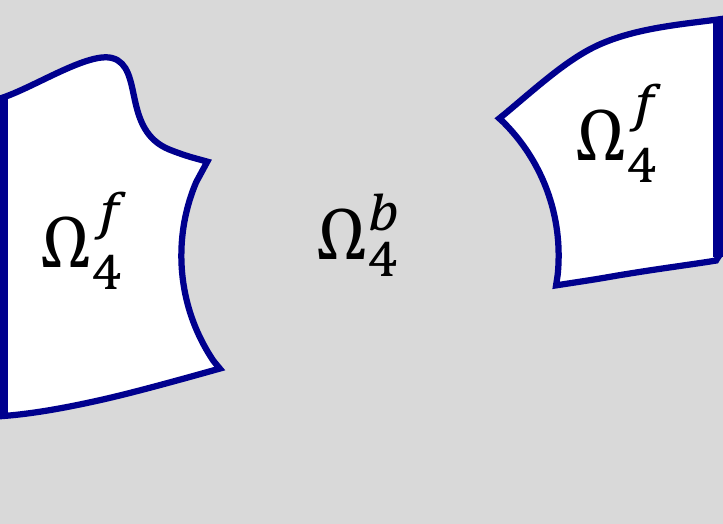}}
		\caption{(a) Example of a ground-truth class map $\mathbf{C}_{gt}$. (b,c) Corresponding foreground~(${ \Omega  }^{f}_{ c }$, white area) and background~(${ \Omega  }^{b}_{ c }$, shaded area) regions where $c=1$ and $c=4$, respectively. Solid blue line is the boundary of the object.}
		\label{FIG:region}
	\end{figure}

	\noindent $\bullet$ \textbf{Boundary-based evaluation.} \quad 
	Meanwhile, many studies~\cite{csurka2013good,perazzi2016benchmark} have pointed out that the prediction performance on the boundaries of objects has a significant impact on the perceived segmentation quality. 
	In fact, the object boundary has higher prediction difficulty than the internal area of the object -- at least two classes are adjacent around the objects' boundaries, which makes class prediction more difficult and ambiguous.
	However, \textit{IoU} equally evaluates all pixels in the image, making it hard to evaluate performances around object boundaries.
	
	{A method~\cite{huttenlocher1992multi} has adopted Hausdorff distance for edge evaluations. This distance measures the greatest of all the distances from a point in one set~(i.e. ground-truth edge points set $A$) to the closest point in the other set~(i.e. estimated edge points set $B$) by} 
	{
		\begin{equation}
			d_H(A, B) = \max \left\{ \sup_{a \in A} \inf_{b \in B} \|a - b\|, \sup_{b \in B} \inf_{a \in A} \|b - a\| \right\},
		\end{equation}
		where $\sup$ and $\inf$ are the supremum and infimum operators. This distance metric can measure similarity between two edge sets without any hyper-parameters, but it is highly sensitive to outliers or noise. In addition, it does not calculate a similarity but a distance which lies on $\left[0,\infty \right]$.
	}

	\begin{figure*}[t]
		\centering
		\subfigure[$\bar{\mathbf{D}}_{1}\left( p \right)$]{\includegraphics[width=0.39\columnwidth]{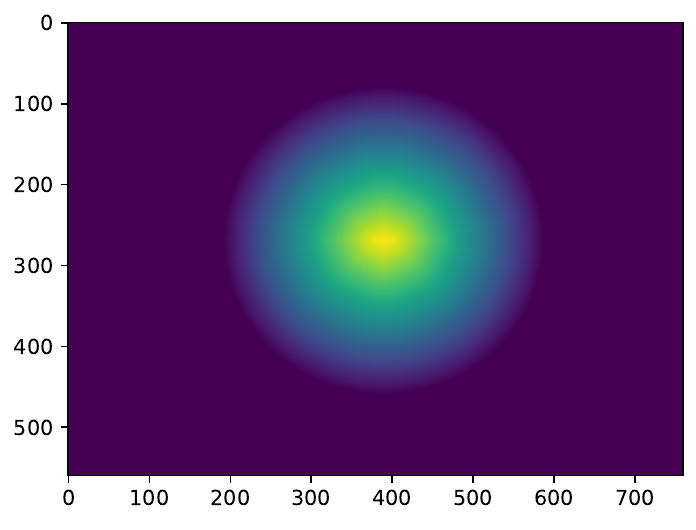}}
		\subfigure[$\bar{\mathbf{D}}_{2}\left( p \right)$]{\includegraphics[width=0.39\columnwidth]{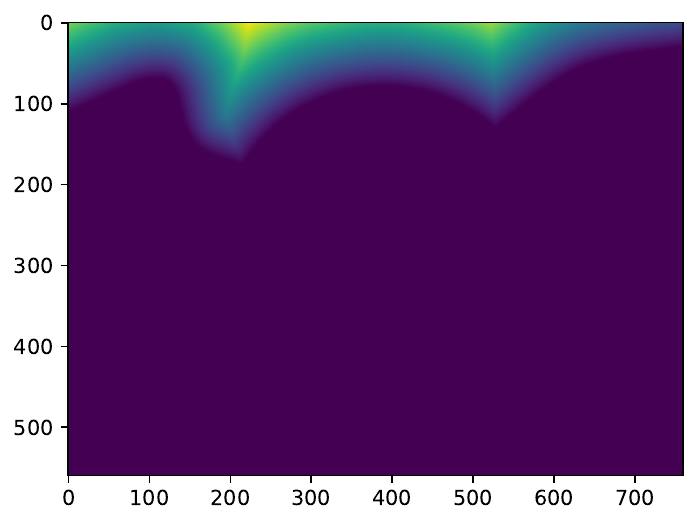}}
		\subfigure[$\bar{\mathbf{D}}_{3}\left( p \right)$]{\includegraphics[width=0.39\columnwidth]{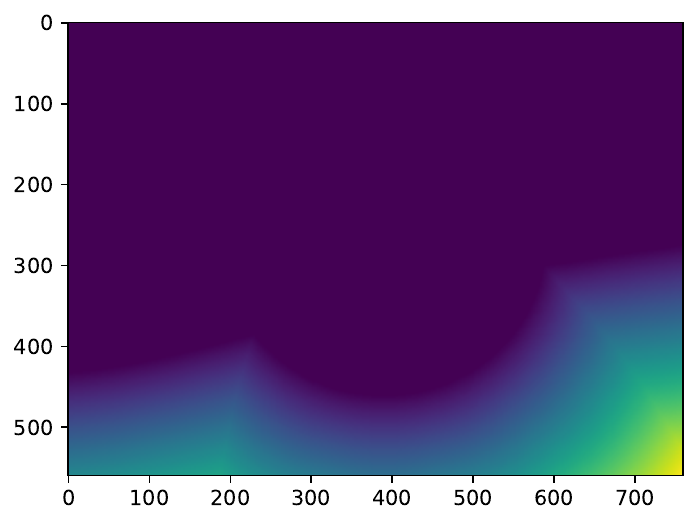}}
		\subfigure[$\bar{\mathbf{D}}_{4}\left( p \right)$]{\includegraphics[width=0.39\columnwidth]{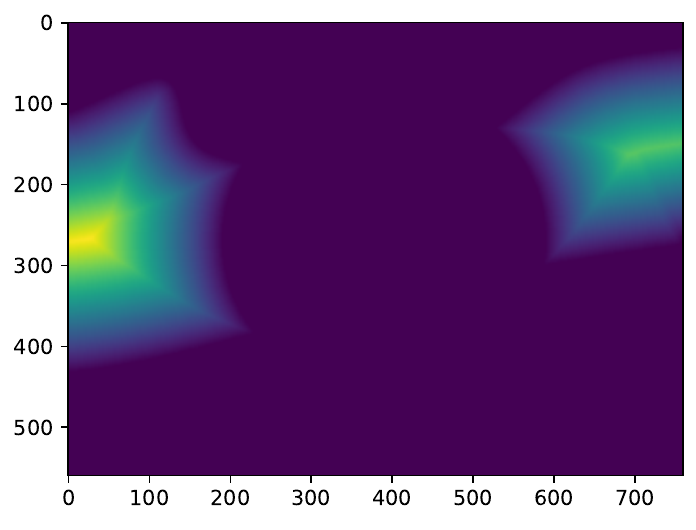}}
		\subfigure[$\bar{\mathbf{D}}\left( p \right)$]{\includegraphics[width=0.46\columnwidth]{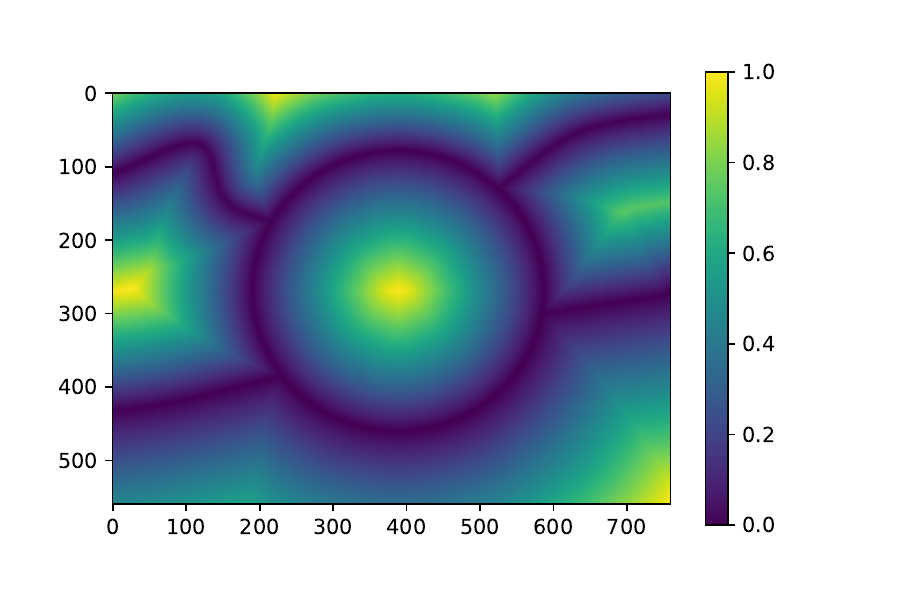}}
		\caption{Examples of distance maps. The ground-truth class map is illustrated in Figure~\ref{FIG:region}~(a).}
		\label{FIG:distance_map}
		\vspace{-10pt}
	\end{figure*}

	Some studies~\cite{martin2004learning, perazzi2016benchmark} evaluated the edge prediction results by measuring edge-based $F_1$ score as in Eq.~\ref{eq:04} only around the objects' edges via bipartite matching. 
	To make the evaluation metric robust to small errors, they set the distance tolerance ($\theta$) for performing bipartite matching between predicted edge points and ground-truth edge points.
	In addition, BF score~\cite{csurka2013good} and Boundary Jaccard~(BJ)~\cite{fernandez2018new} were also proposed to capture the edge prediction results. 
	
	The evaluation metrics can measure the boundary prediction performance; however they still require certain distance tolerance values~($\theta$), which determine the boundary region in the scene. 
	Selecting a distance tolerance~($\theta$) leads an inflexibility in evaluation measures.
	Furthermore, region-based evaluation metrics and boundary-based evaluated metrics are separate, making it difficult to compare the image segmentation performances at a glance.
	In this work, we propose a single evaluation metric that can measure both evaluation factors~(i.e., region- and boundary-based) in Section.~\ref{sec:proposed}.
	\vspace{-10pt}

	\section{Proposed Evaluation Metric}
	\label{sec:proposed}
	\vspace{-5pt}
	
	\subsection{Motivation and main ideas}
	\label{sec:mot}
	
	We observed that it is more difficult to infer the label of the boundary of each object.
	This is because the inference probability of each label is mixed at the boundary.
	Unfortunately, the details of an object are usually at the boundary.
	Even if most of the object region is predicted correctly, if it fails to predict the boundary of the object, it looks qualitatively bad. 	
	In addition, missing object details can lead to poor results in applications of semantic segmentation.
	
	To evaluate segmentation results while considering these issues, we first generate a weighted map based on a distance transform in Section.~\ref{sec:weighted}. 
	The weight map emphasizes the importance of the object's boundaries.
	We then calculate the weighted intersection over union (\textit{wIoU}) using the generated weight map in Section.~\ref{sec:prop2}.

	\subsection{Weighted map generation}
	\label{sec:weighted}

	The problem of a distance transform is to compute the distance of each point from the boundary of foreground region~\cite{fabbri20082d}.
	First, it assigns fore- and back-ground regions according to the class label $c$ as
	\begin{equation}
		{ \Omega  }^{f}_{ c } =\left\{ { p }|{ \mathbf{C}_{ gt }\left( p \right) =c } \right\},
		{ \Omega  }^{b}_{ c } =\left\{ { q }|{ \mathbf{C}_{ gt }\left( q \right) \neq c } \right\}.
	\end{equation}
	Each region consists of a set of points. 
	Figure~\ref{FIG:region} shows predicted map~($\mathbf{C}$)~(Figure.~\ref{FIG:region} (a)) and its corresponding fore- and back-ground regions~(${ \Omega}^{f}_{ c },{ \Omega  }^{b}_{ c }$) where $c=1$ and $c=4$ (Figure.~\ref{FIG:region} (b) and (c)). Solid blue lines are the boundaries of objects.
	
	
	Next, it generates a distance map that consists of a distance value of each point~$p$ by computing the smallest distance from the back-ground region~${ \Omega  }^{b}_{ c }$. It is represented by	
	\begin{equation}
		\mathbf{D}_{c}\left( p \right)  :=  \mathrm{min}\left\{ \mathrm{d}\left( p,q \right)  |  p \in { \Omega  }^{f}_{ c }, q \in { \Omega  }^{b}_{ c } \right\},  
	\end{equation}
	where $\mathrm{d}\left( p,q \right)$ is the pixel distance between two points $p$ and $q$.
	To compute the distance, we can adopted $\rho $-norm (also called $l_\rho$ norm) defined by
	$\mathrm{d}\left( p,q \right)  = { \left( { \left| p_x - q_x \right|  }^{ \rho  }+{ \left| p_y - q_y \right|  }^{ \rho  } \right)  }^{  1 / \rho    },$
	where $\rho\geq1$.
	$\rho$-norm generates various distance metrics such as Manhattan~($ \rho=1$), Euclidean~($\rho=2$), and Chessboard~($\rho=\infty$) distances according to its value $\rho$.
	In this work, we set $\rho$ as $2$ to follow Euclidean distance for generating the weight map~$\mathbf{D}_{c}$.
	As it mentioned in Section~\ref{sec:mot}, we aim to generate a weight map according to the distance from the object boundary.
	Calculating Euclidean distances can efficiently approximate to compute the normal distances from the boundaries.
	As a result, the distance maps for all categories $\mathcal{C}$ can be simply calculated.

\begin{figure}[]
		\centering
		\subfigure[Ground truth $\mathbf{C}_{gt}$]        {\includegraphics[width=1\columnwidth]{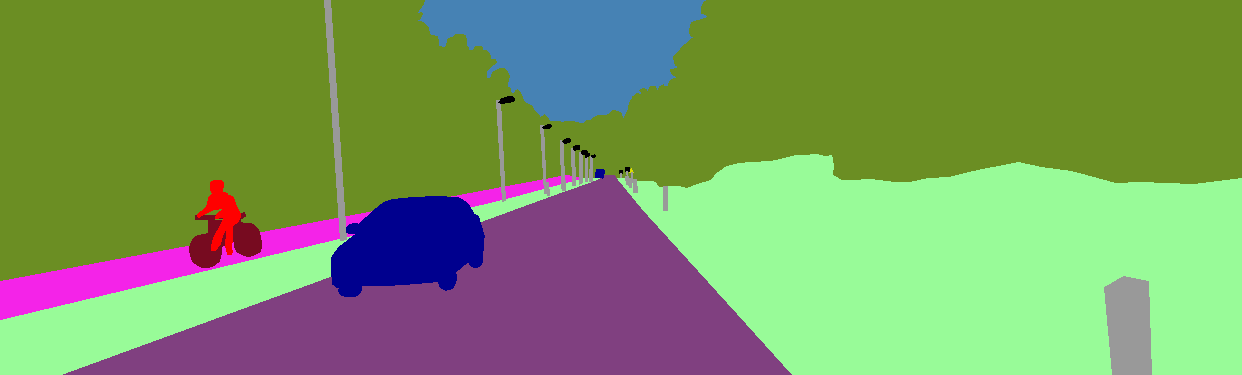}}
		\subfigure[Weight map $\mathbf{W}$, $\alpha=0.01$]{\includegraphics[width=1\columnwidth]{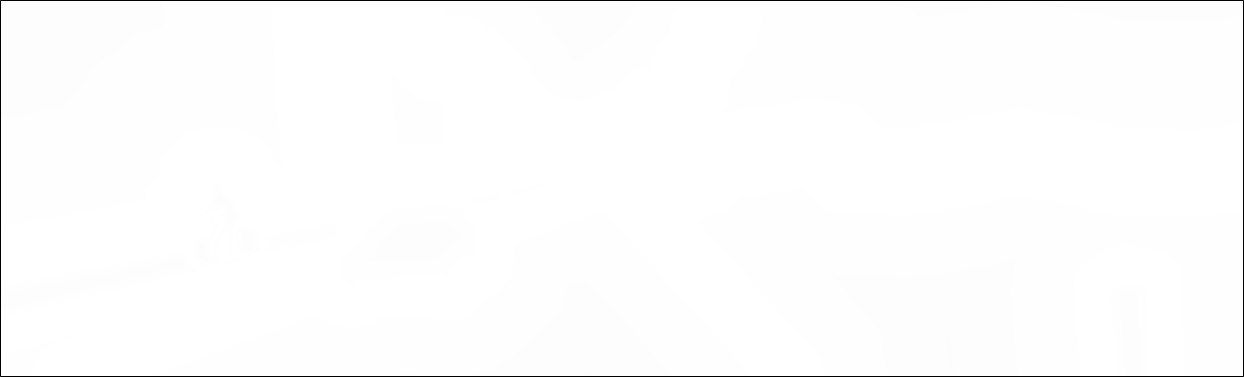}}
		\subfigure[Weight map $\mathbf{W}$, $\alpha=0.1$] {\includegraphics[width=1\columnwidth]{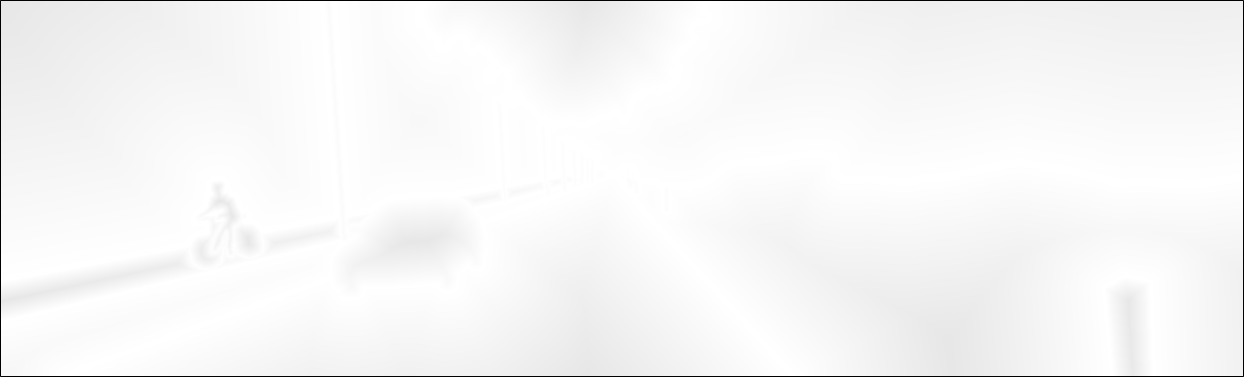}}
		\subfigure[Weight map $\mathbf{W}$, $\alpha=1$]   {\includegraphics[width=1\columnwidth]{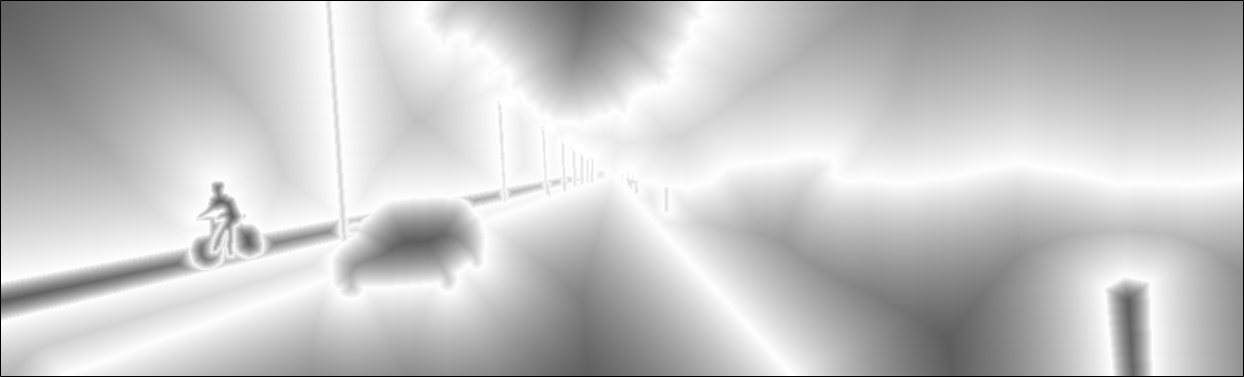}}
		\subfigure[Weight map $\mathbf{W}$, $\alpha=10$]  {\includegraphics[width=1\columnwidth]{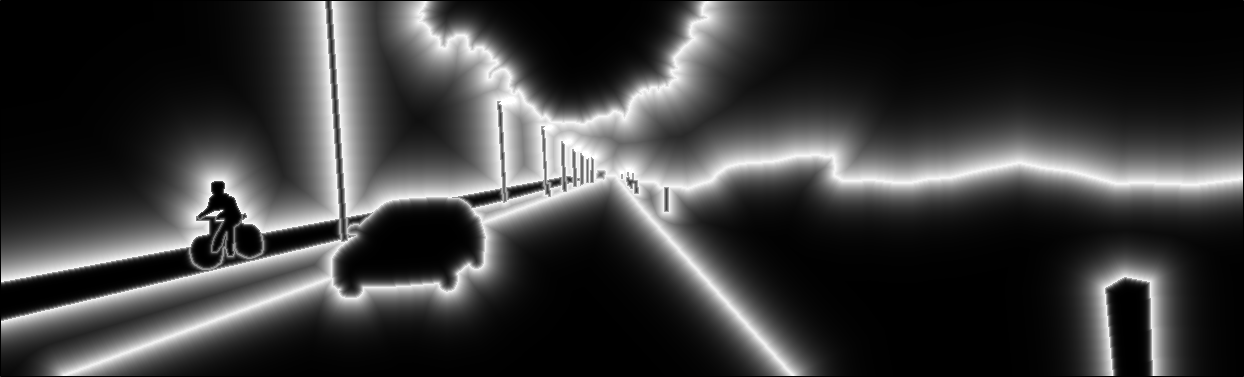}}
		\subfigure[Weight map $\mathbf{W}$, $\alpha=100$] {\includegraphics[width=1\columnwidth]{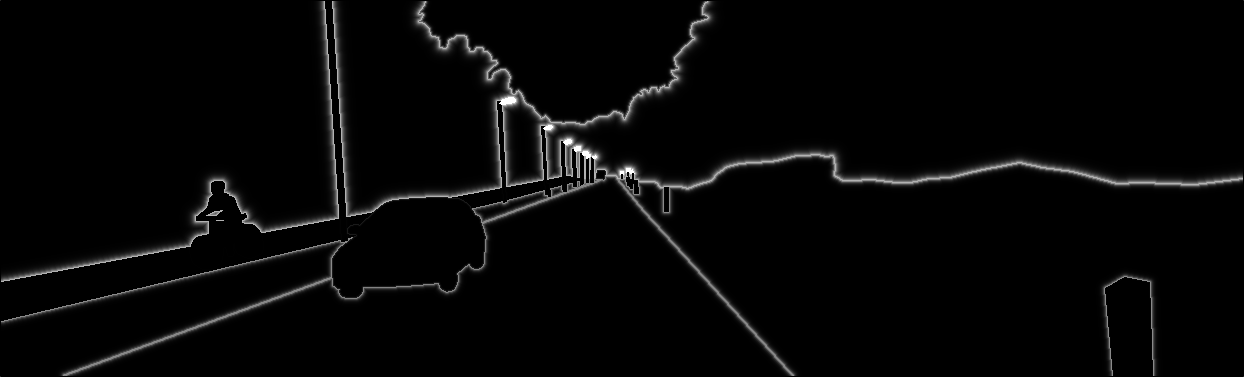}}
		\caption{Ground truth and its weight maps according to the boundary importance factor $\alpha$. The range of weight map is $[0,1]$. White and black pixels denote weight values $1,0$, respectively. A small boundary importance factor leads a uniform weight map. On the other hand, a large boundary importance factor leads high weights around the boundaries of the objects and regions.}
		\label{FIG:maps}
	\end{figure}

	However, the distance map of each object has different distance range due to the difference in size between the objects. 
	It causes unfair performance evaluation in many cases.
	To handle this, we designed the regularized distance as
	\begin{equation}
		\bar{\mathbf{D}}_{c}\left( p \right) = \frac{\mathbf{D}_{c}\left( p \right)}{\mathrm{max} \big(\mathbf{D}_{c}\left( p \right)\big)},
		\label{eq:regul}
	\end{equation}
	Then it lies on $[0,1]$ for all pixel points $p$ and classes $c$.
	Note that if a class $c$ has multiple instances, the regularization in Eq.~\ref{eq:regul} is performed for each instance.
	Figure.~\ref{FIG:distance_map} shows examples of regularized distances.
	Considering all the class in the class map $\mathbf{C}$, the distance maps $\bar{\mathbf{D}}_{c}$ can be combined into one distance map as follows
	\begin{equation}
		\bar{\mathbf{D}}\left( p \right) = \left\{\bar{\mathbf{D}}_{c}\left( p \right) | 1 \leq  c \leq \mathcal{C} \right\},
	\end{equation}
	An example of the combined distance map is shown in Figure.~\ref{FIG:distance_map}~(e).
	Finally, {we designed a weight map which is an exponential decay function with different $\alpha$ values by} 
	\begin{equation}
		\mathbf{W}\left( p \right) = e^{-\alpha\bar{\mathbf{D}}\left( p \right)},  
		\label{eq:weight_map}
	\end{equation}
	where $\alpha>0$ is a boundary importance factor. 
	As shown in Figure.~\ref{FIG:maps}, the value of $\alpha$ determines the importance of boundaries in the scene.
	When the boundary importance is small~($\alpha=0.01$), the weight map is uniformly distributed~(Figure.~\ref{FIG:maps} (b)). It means that all areas have about the same weights.
	On the other hand, when the boundary importance is comparatively large
	~($\alpha=100$), it emphasizes boundaries in the scene~(Figure.~\ref{FIG:maps} (f)).
	In this way, we can set the weight map very simply by manipulating only one parameter $\alpha$.

	\subsection{Weighted intersection over union}
	\label{sec:prop2}
	
	Based on the calculated weight map in Section.~\ref{sec:weighted}, we propose a new evaluation metric called Weighted Intersection over Union~(\textit{wIoU}) by
	\begin{equation}
		wIoU=\frac { \left| \mathbf{C}\cap (\mathbf{ C }_{ gt }\circ \mathbf{W}) \right|  }{ \left| \mathbf{C}\cup (\mathbf{ C }_{ gt }\circ \mathbf{W}) \right|  }, 
		\label{eq:wiou}
	\end{equation}
	where $\circ$ is an element-wise product called Hadamard product~\cite{horn1990hadamard}.
	In this step, we omit $p$ for convenience and all terms in the Eq.~\ref{eq:wiou} have the same dimension~(i.e., size of an input image).
	Comparing the proposed \textit{wIoU} with conventional \textit{IoU}~(Eq.~\ref{eq:iou}), it emphasizes the importance of each pixel based on the weight map~$\mathbf{W}$.
	{The overall procedure of calculating wIoU score is described in Algorithm.~\ref{alg1}.}
	{In addition, the codes are available on \\ \url{https://github.com/engzenia/wIoU}}
	
	\vspace{-10pt}

	\begin{algorithm}[t]
		\footnotesize

		\KwIn{Estimated class map $\mathbf{C}$, Ground-truth class map $\mathbf{C}_{gt}$}
		\KwOut{Weighted intersection over union score $wIoU$}
		
		\For{each class label $c$ in $\mathbf{C}_{gt}$}{

			\For{pixel index $p,q$ in $\mathbf{C}_{gt}$}{
				$\left({ \Omega  }^{f}_{ c }, { \Omega  }^{b}_{ c }\right) = $ Get fore- and back-ground regions$\left(c,\mathbf{C}_{gt}\right);$
			}
			
			\For{pixel index $p$ in ${ \Omega  }^{f}_{ c }$, pixel index $q$ in ${ \Omega  }^{b}_{ c }$}{
				$\mathbf{D}_{c}\left( p \right) = $ Distance map generation$\left({ \Omega  }^{f}_{ c }, { \Omega  }^{b}_{ c }\right);$
			}
			
			\For{pixel index $p$ in $\mathbf{D}_{c}\left( p \right)$}{
				\textcolor[rgb]{0,0.5,0.2}{\% distance map regularization} \\
				$\bar{\mathbf{D}}_{c}\left( p \right) = \dfrac{\mathbf{D}_{c}\left( p \right)}{\mathrm{max} \big(\mathbf{D}_{c}\left( p \right)\big)};$ 
			}	
		}
		
		$\bar{\mathbf{D}}\left( p \right) = \left\{\bar{\mathbf{D}}_{c}\left( p \right) | 1 \leq  c \leq \mathcal{C} \right\};$ \textcolor[rgb]{0,0.5,0.2}{\% a combined distance map}
		
		\For{pixel index $p$ in $\bar{\mathbf{D}}\left( p \right)$}{
			$\mathbf{W}\left( p \right) = \exp(-\alpha\bar{\mathbf{D}}\left( p \right));$ \textcolor[rgb]{0,0.5,0.2}{\% weight map calculation}
			
			$wIoU= \dfrac{ \left| \mathbf{C}\cap (\mathbf{ C }_{ gt }\circ \mathbf{W}) \right|  }{ \left| \mathbf{C}\cup (\mathbf{ C }_{ gt }\circ \mathbf{W}) \right|  };$
		}
		\caption{{An algorithm of the propose wIoU}}
		\label{alg1}
	\end{algorithm}

	\begin{figure}[]
		\centering
		\subfigure[Scene \#01]{\includegraphics[width=0.32\columnwidth]{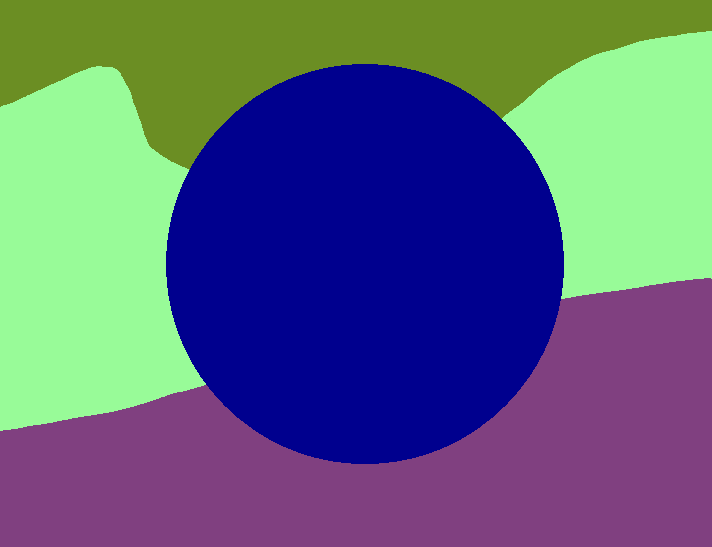}}
		\subfigure[Scene \#02]{\includegraphics[width=0.32\columnwidth]{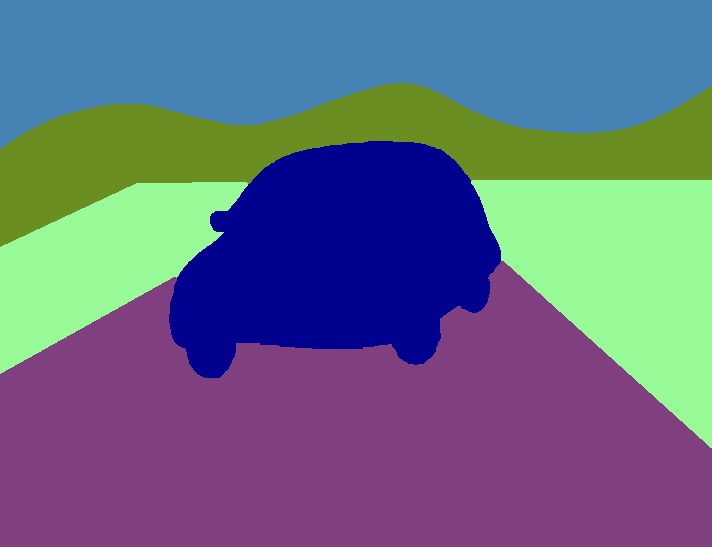}}
		\subfigure[Scene \#03]{\includegraphics[width=0.32\columnwidth]{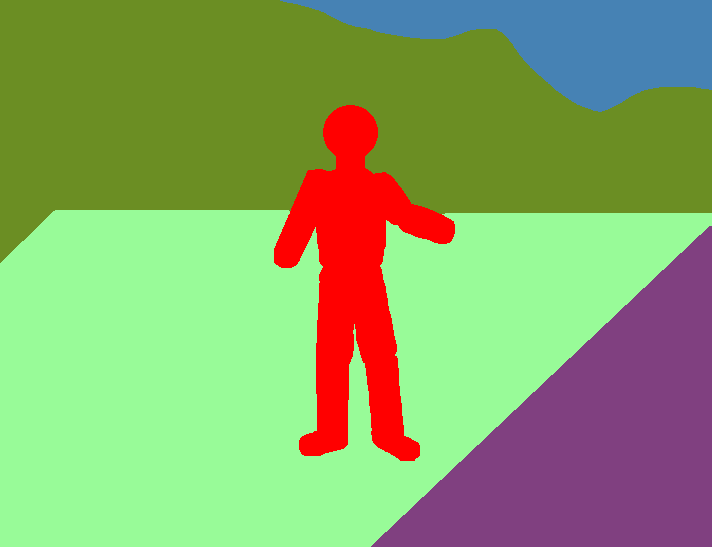}}
		\caption{Examples of synthetic test scenes and their segments}
		\label{FIG:test_ex_scene}
		\vspace{-5pt}
	\end{figure}

	\begin{figure*}[t]
		\centering	
		\subfigure[Erode lv5]{\includegraphics    [width=0.29\columnwidth]{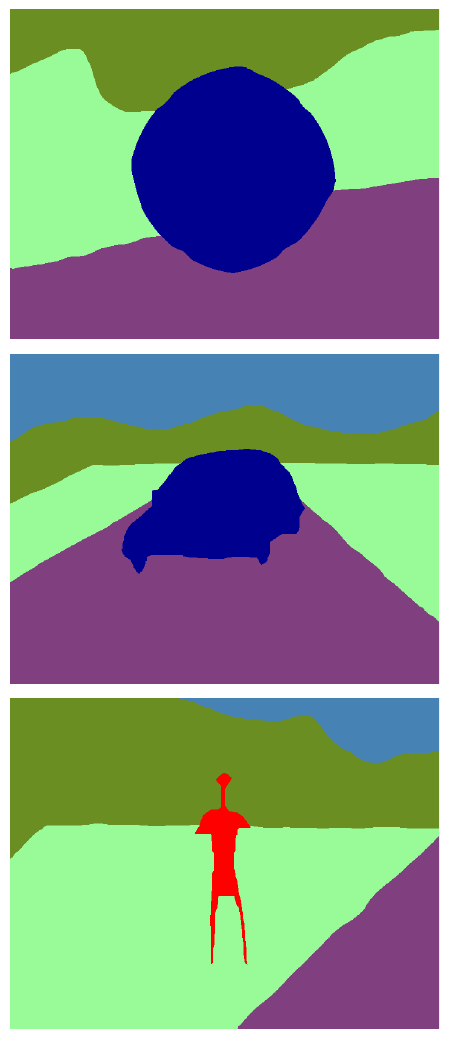}}
		\subfigure[Erode lv3]{\includegraphics    [width=0.29\columnwidth]{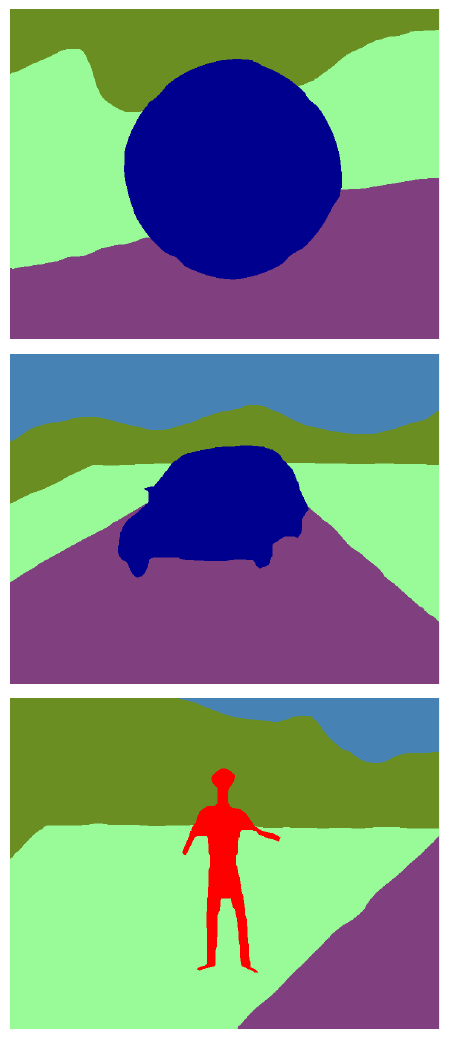}}
		\subfigure[Erode lv1]{\includegraphics    [width=0.29\columnwidth]{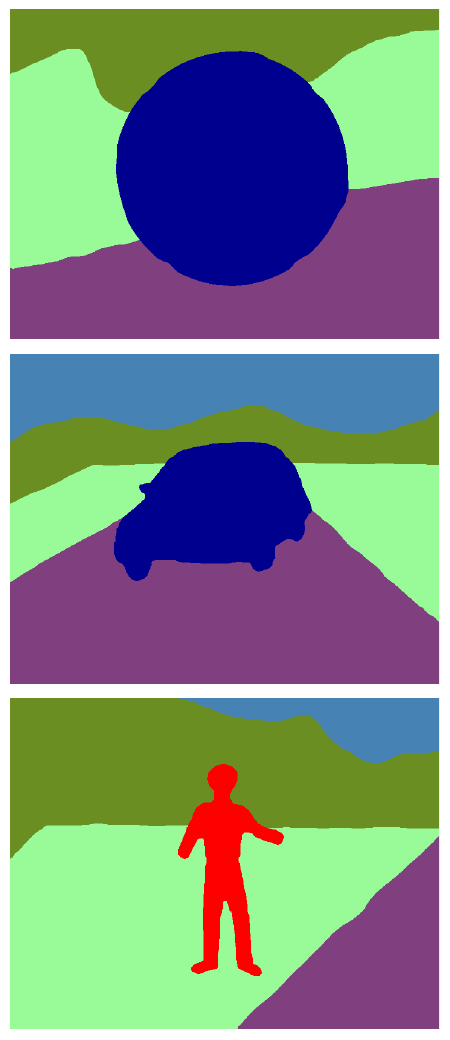}}
		\subfigure[Test segments]{\includegraphics[width=0.29\columnwidth]{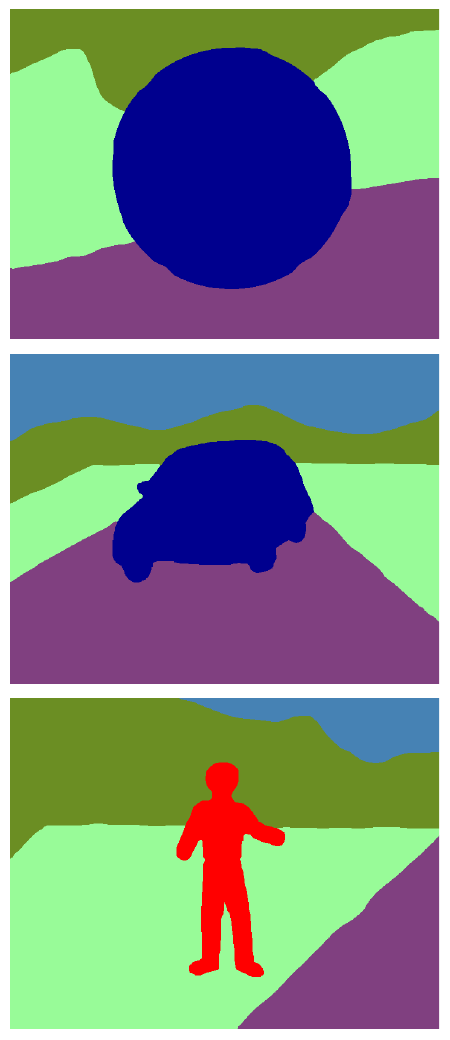}}
		\subfigure[Dilate lv1]{\includegraphics   [width=0.29\columnwidth]{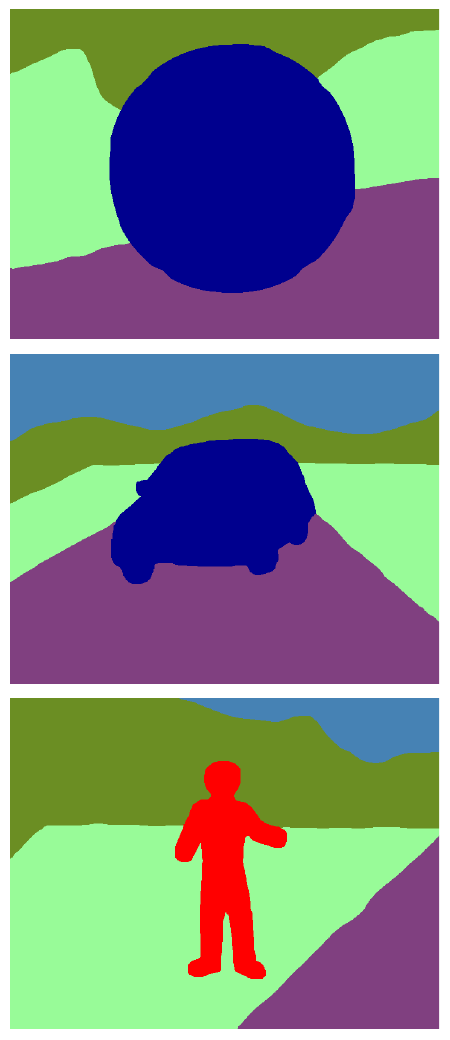}}
		\subfigure[Dilate lv3]{\includegraphics   [width=0.29\columnwidth]{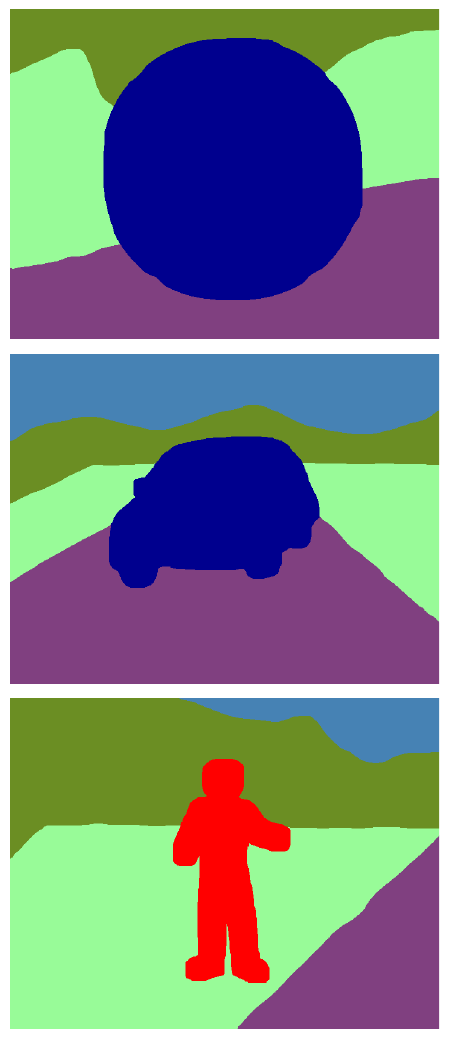}}
		\subfigure[Dilate lv5]{\includegraphics   [width=0.29\columnwidth]{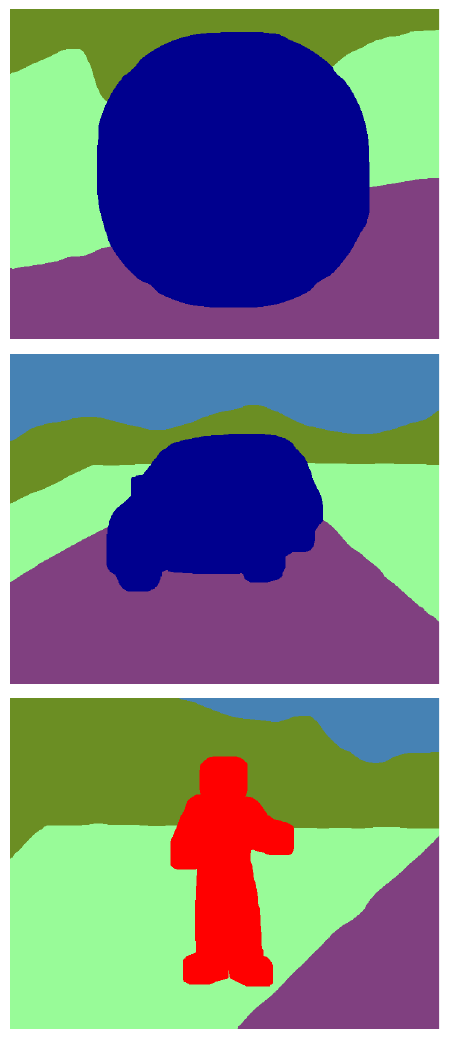}}		
		
		\caption{Examples of generated test segments. (d) Generated test segments for each scene including prediction error. Based on morphological image processing~(5 different levels of erosion~(a--c) and dilation~(e--g)), the test segments were augmented.}
		\label{FIG:test_ex}
		\vspace{-10pt}
	\end{figure*}

	\section{Experimental Results}
	\label{sec:exp}
	
	\subsection{Datasets}
	
	To compare evaluation metrics, we built a segment dataset as shown in Figure.~\ref{FIG:test_ex_scene}.
	We followed the \texttt{KITTI} \cite{Geiger2012CVPR} color assignments for classes. At the center of each scene, we placed an object~(e.g., \tb{car}, \tr{person}). 
	For each scene, we created test segments (Figure.~\ref{FIG:test_ex}~(d)) that include prediction errors around boundaries, which commonly occur. To consider more possible situations, we applied morphological image processing to the objects in test images. 
	We performed 5 different levels of erosion and dilation for each image.
	As a result, the total number of segment images is $33$.
	These test images will be used to compare overall similarities between different evaluation metrics in Section.~\ref{exp:comp}.
	
	Additionally, we created three test segment images that contain the same number of error pixels~(FP+FN), but the positions of the error pixels are different, as illustrated in Figure.~\ref{FIG:unusual}. These test scenarios are quite unusual in semantic segmentation.
	However, we can precisely compare how each evaluation metric works under the same number of error pixels.

	\subsection{Comparisons of different metrics}
	\label{exp:comp}
	
\begin{figure}[t]
		\centering
		\includegraphics[width=1\columnwidth]{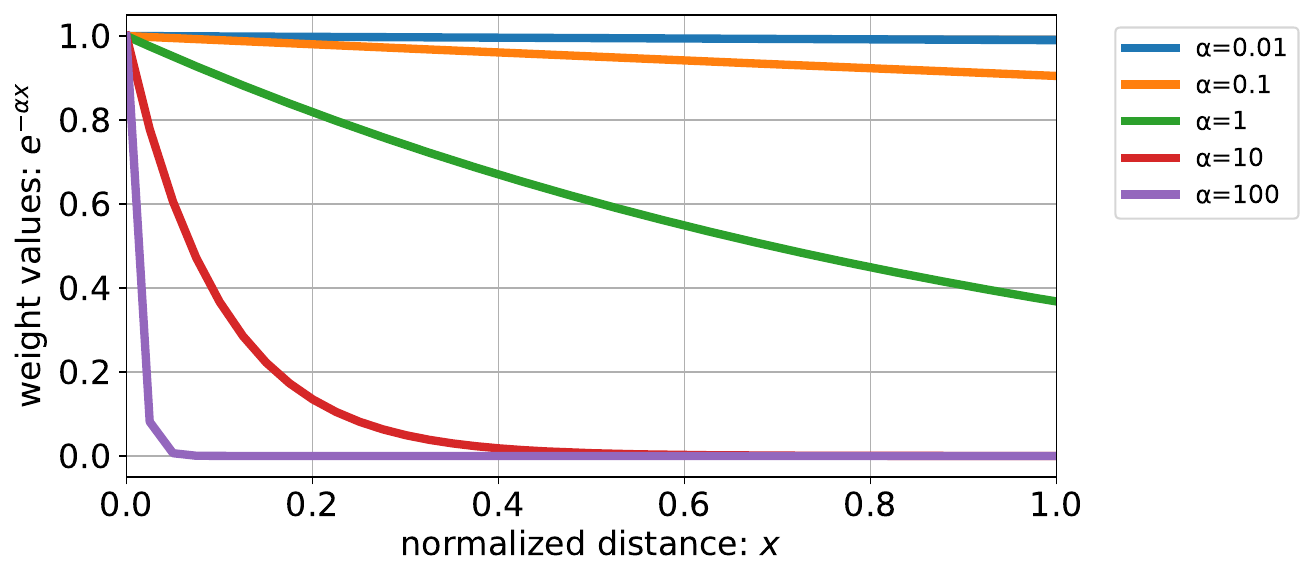}
		\caption{{Weight values $e^{-\alpha x}$ with different $\alpha$ values. $\alpha$ follows logarithmic scale with a base of 10. When the $x$ value is close to $0$ and $1$, it indicates the boundary and the innermost area of the object, as described in Eq.~\ref{eq:weight_map}.}}
		\label{FIG:10}
		\vspace{-10pt}
	\end{figure}

	In this section, we compare three different evaluation metrics: 1) Intersection over union~(\textit{IoU}), 2) Proposed weighted IoU~(\textit{wIoU}), 3) edge-based score $F_1$~\cite{martin2004learning}. 
	For the proposed \textit{wIoU}, we tested five different levels of boundary importance factors $\alpha$ as $0.01$, $0.1$, $1$, $10$ and $100$.
	{These values follow a simple logarithmic scale with a base of 10, but can create distinctive functions as shown in Fig.~\ref{FIG:10}. 
		The upper two values $(10, 100)$ assign similar importance across the overall area ($0 \leq x \leq 1$), while the lower two values $(0.01, 0.1)$ only emphasize the areas around the boundaries ($x \simeq 0$).
		Furthermore, as shown in Fig.~\ref{FIG:maps}, the five alpha values effectively highlight the importance of boundaries or areas with different criteria.}
	
	For the edge-based score $F_1$, we set the distance tolerance value $\theta$ as 3 pixels.
	To compare the evaluation results of the metrics, correlations $\rho_{X,Y}$ and sum of difference $D_{X,Y}$ are calculated as follows:
	\begin{equation}	
		\rho_{X,Y} = \frac{cov(X,Y)}{\sigma_{X}\sigma_{Y}},
	\end{equation}
	\begin{equation}
		D_{X,Y} = E[|X-Y|],
	\end{equation}
	where $X$, $Y$ are sets of evaluation values calculated by each evaluation metric,
	$cov(X,Y)= E[(X-\mu_X)(Y-\mu_Y)]$ is a covariance matrix between $X$ and $Y$. $\mu$ and $\sigma$ are mean and standard deviations of the evaluation values, respectively.
	
	\begin{figure}[t]
		\centering
		\subfigure[Correlations]			 {\includegraphics[width=0.95\columnwidth]{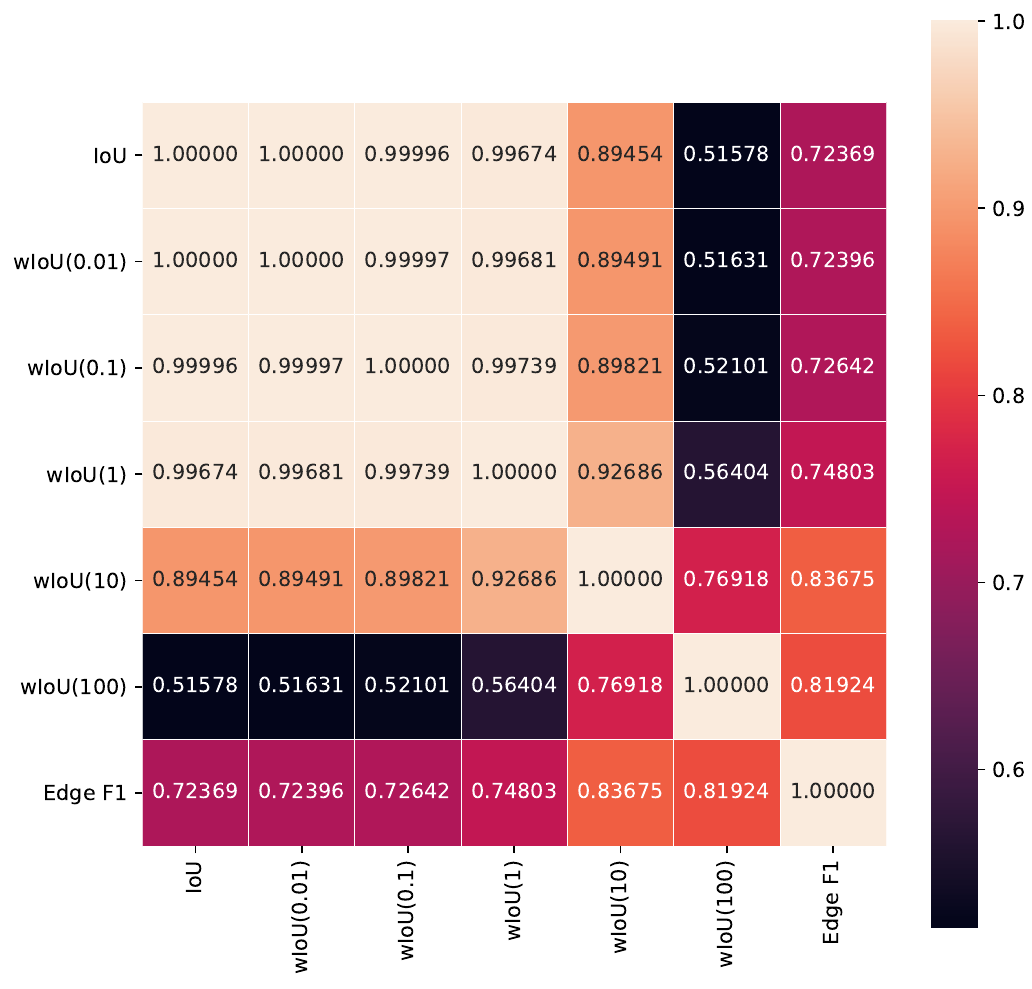}}
		\subfigure[Summations of differences]{\includegraphics[width=0.95\columnwidth]{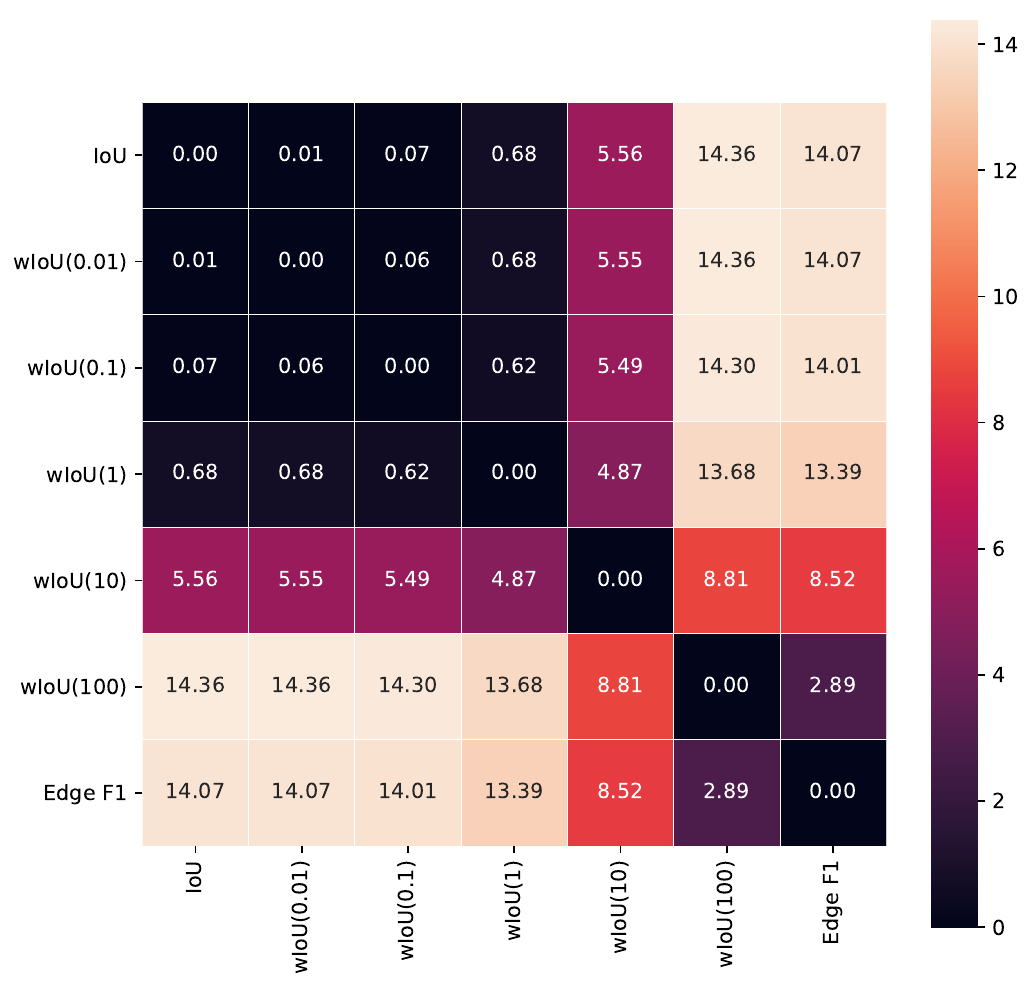}}
		\caption{Comparison matrices of different metrics}
		\label{FIG:corr}
		\vspace{-8pt}
	\end{figure}

	The correlation $\rho_{X,Y}$ measures strength of a relationship between two variable sets~$X,Y$.
	Thus, although the absolute values between two variable sets are different, they have a correlation value when the tendency of value change is similar. The value of correlation lies on $[0,1]$, and the value $1$ indicates the high similarity.
	On the other hand, sum of difference $D_{X,Y}$ measures the absolute value difference between the two variable sets.
	It can lie on $[0,\infty]$, and the value $0$ indicate the high similarity.
	Based on both evaluations, we can measure a relationship and differences between metrics.
	
	Figure.~\ref{FIG:corr} illustrates correlations and summations of differences between \textit{IoU}, \textit{wIoU}, and edge-based $F_1$.
	\textit{wIoU}s with low boundary importance factors ($\alpha=0.01, 0.1$) are very similar with \textit{IoU} in terms of both correlation and sum of differences. 
	On the other hand, \textit{wIoU}s with high boundary importance factors ($\alpha=10, 100$) show the high similarity with edge-based $F_1$.
	The sum of differences between edge-based $F_1$ and \textit{wIoU} series become lower as the boundary importance factor increases.
	In this way, the proposed \textit{wIoU} combines both conventional region and boundary-based evaluation metrics depending on $\alpha$.
	
\begin{figure}[]
		\centering
		\includegraphics[width=1\columnwidth]{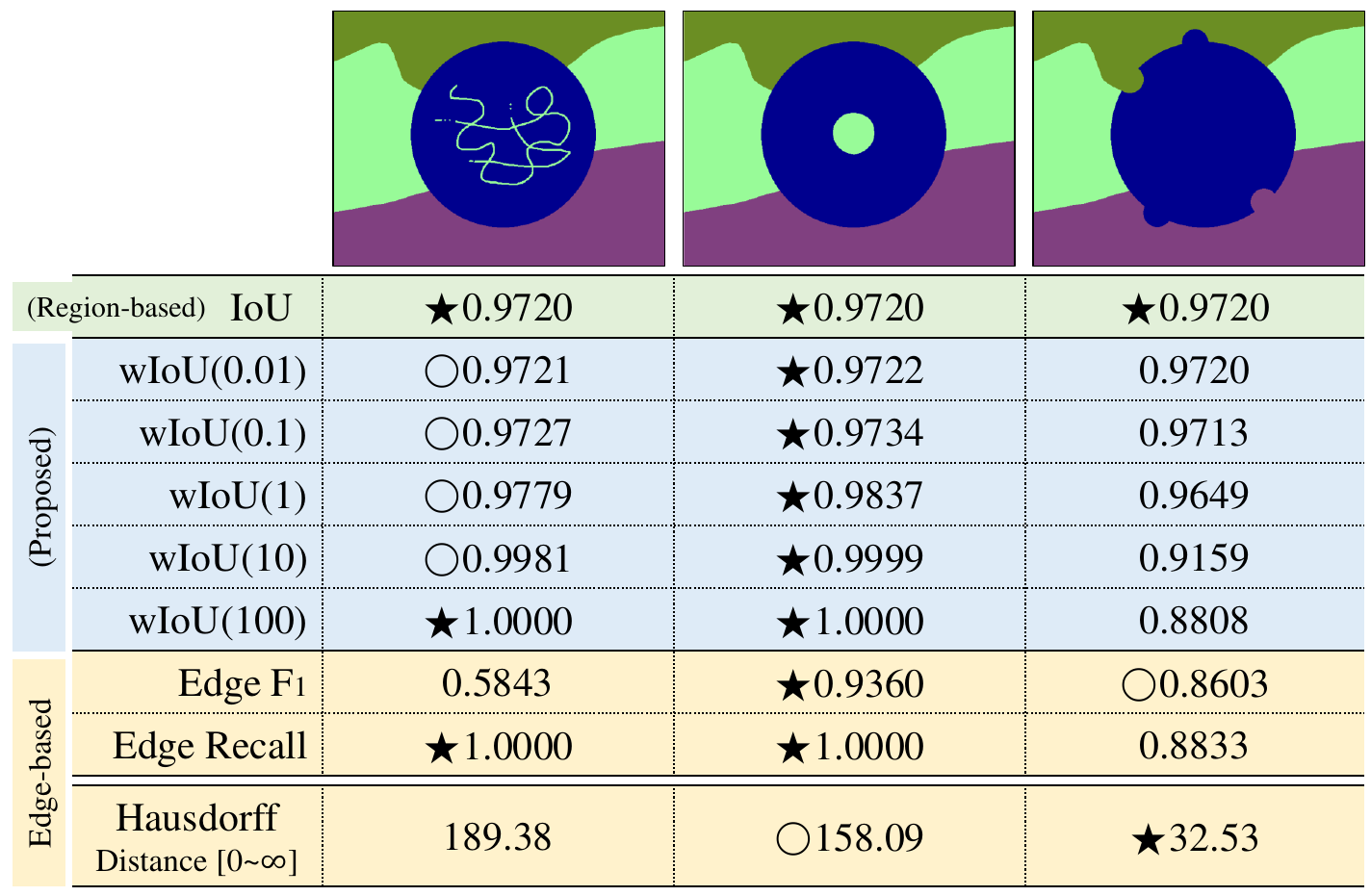}
		\caption{Evaluations of different metrics on specific test images. {The ground-truth is shown in Fig.~\ref{FIG:test_ex_scene} (a). $\bigstar$ and $\bigcirc$ indicate the best and second best results.}}
		\label{FIG:unusual}
		\vspace{-8pt}
	\end{figure}

	For further comparisons of evaluation metrics, we additionally tested three test images as illustrated in Figure.~\ref{FIG:unusual}. Those test images were generated from Scene \#01~(Figure.~\ref{FIG:test_ex_scene} (a)) while they keeps the same numbers of error pixels~(FP+FN).
	Therefore region-based \textit{IoU} metric showed the same evaluation results~($0.972$) for all test images.
	On the other hand, other evaluation metrics~(i.e., \textit{wIou}, edge-based $F_1$, edge-based recall) measure different aspects of performance evaluations.
	\textit{wIoU} with small boundary importance factors~($\alpha=0.01,0.1$) are very relevant to $IoU$ metric as they worked in common test images in Figure~\ref{FIG:corr}. 
	When boundary importance factor was set large~($\alpha=100$) for \textit{wIoU}, it works almost the same with edge-based Recall~($\frac{TP}{TP+FN}$).
	This result indicates that \textit{wIoU} with large boundary importance is difficult to measure false positive~(FP) pixels which are not around ground-truth boundaries.
	\textit{wIoU} with a boundary importance factor~($\alpha=1$) makes difference between the test images {compared to \textit{IoU}}. It gives more weight to the boundaries, but it can evaluate both the regions and the boundaries in a reasonable way.
	
	An edge-based $F_1$ was highly dependent on the predicted boundary in the scene; it evaluated very different values for each test image~($0.5843,0.936,0.8603$) despite of the same error pixels.
	{We also tested the Hausdorff distance for these test scenarios. It computed distance values of ($189.38, 158.09, 32.53$), where smaller values indicate higher similarity. It ranked the third image as the best segmentation result, with a large gap unlike other edge-based measures. 
		The distance highly focuses on measuring the localization error of the predicted boundaries. 
		It calculates a large error if false positives are spatially far from the actual boundary.
		This means that the measure can be sensitive to outliers and noisy boundaries. Furthermore, it requires an additional scale factor to normalize the distance value from $\left[0, \infty\right]$ into a similarity value $\left[0, 1\right]$.}
	
\begin{figure*}[t]
		\centering
		\includegraphics[width=2\columnwidth]{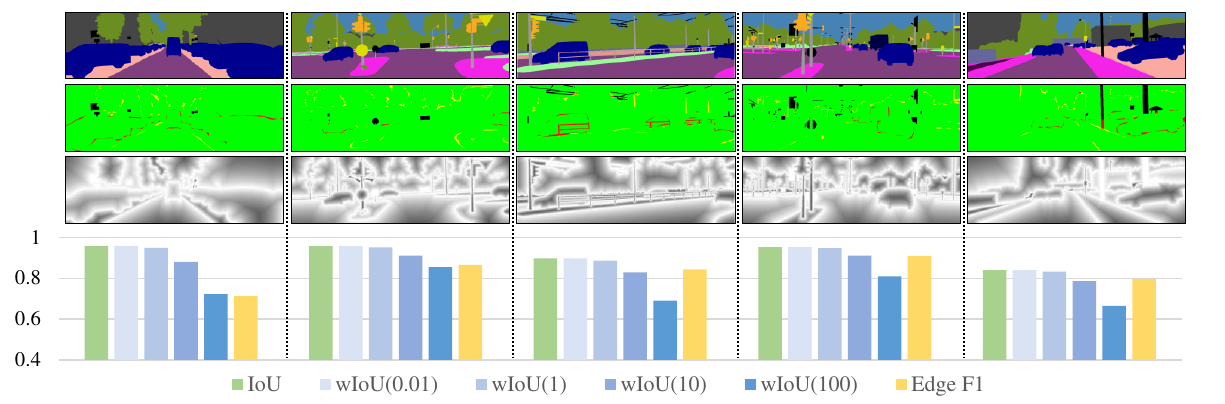}
		\caption{{Comparison of evaluation results based on different metrics in real scenes. First and second rows are ground-truth segments, and error map of predicted segmentation results. Third row images are weighted map of \textit{wIoU}($\alpha=1$). The last row illustrates evaluation results based on each metric.}}
		\label{FIG:real_data}
		\vspace{-5pt}
	\end{figure*}

	{In real-world scenarios, Fig.~\ref{FIG:real_data} shows the comparison of evaluation metrics of different test frames.
		The test frames contain various objects and background classes which are much more complex and general than synthesized test scenarios in Fig.~\ref{FIG:test_ex}. 
		The results demonstrate that the proposed \textit{wIoU} series effectively measures segmentation performance based on the importance of both boundary and region. When the boundary importance factor is high $(\alpha = 100)$, \textit{wIoU} calculates a lower performance than the edge-based $F_1$ score.
		That is because, while the edge-based metric addresses a binary problem of determining whether a pixel is part of an edge or not, the proposed \textit{wIoU} also requires predicting the class of the edge, making it a much more difficult problem.
		A high alpha value in the \textit{wIoU} metric can result in a much stricter evaluation of edges compared to edge-based $F_1$.
	}
	
	\vspace{-5pt}

	\section{{Discussions and Future Works}}
	\label{sec:discussions} 	
	{
		\textit{wIoU} has several limitations and concerns. First, boundary importance factor $\alpha$ must be determined. We recommend setting $\alpha=1$ to evaluate both region and edge of the segmentation results in a balanced manner. 
		However, the organizer of the competition will determine $\alpha$ in advance, depending on which area~(boundary or region) they consider important.
		Second, in cases where the shape complexity of an object is high, boundaries may have a greater impact on performance evaluation than regions. For example, if the ratio of the object's boundary to its area is high~(see a circle and a human in Fig.~\ref{FIG:test_ex_scene}), the boundary occupies a higher proportion of the total evaluation pixels, and the boundary performance has a significant impact on the final score due to the \textit{wIoU}'s policy. While this may be natural and reasonable, it is an issue that needs further consideration for a fairer evaluation.
	}
	{
		Based on our study, there are a few areas that could be explored for the future works.
		Beyond using it as an evaluation metric, we can adopt the proposed \textit{wIoU} as an optimization loss for deep neural networks.
		To demonstrate the efficiency of \textit{wIoU} in training networks, we have to design various structures of deep neural networks and experimentally demonstrate performance improvements in the next works.
	}

	\section{Conclusions}
	\label{sec:conclusion} 
	{
		In this work, we proposed a new evaluation measure (\textit{wIoU}) for evaluating semantic segmentation. 
		To this end, we calculated boundary distance map, and generated weight maps, allowing weighted evaluation for each pixel based on a boundary importance factor. 
		The proposed \textit{wIoU} can evaluate both boundary and region by setting a boundary importance factor with out any hyper-parameters. 
		We validated the effectiveness of \textit{wIoU} on a dataset of synthetic and real test images. The experimental results demonstrated  flexibility and rationality of the proposed \textit{wIoU} evaluation metric.
	}

	{\small
		\bibliographystyle{ieee}
		\bibliography{egbib}
	}

\end{document}